\documentclass{article} % For LaTeX2e
\usepackage{iclr2026_conference,times}

\usepackage{amsmath,amsfonts,bm}

\def\eqref#1{equation~\ref{#1}}
\def\1{\bm{1}}

\DeclareMathAlphabet{\mathsfit}{\encodingdefault}{\sfdefault}{m}{sl}
\SetMathAlphabet{\mathsfit}{bold}{\encodingdefault}{\sfdefault}{bx}{n}

\usepackage{hyperref}
\usepackage{url}

\usepackage{dsfont}
\usepackage{amsmath}

\usepackage{amssymb}
\usepackage{amsthm}

\usepackage{units}
\usepackage{mathtools}
\usepackage{color}
\usepackage{floatrow}
\newfloatcommand{capbtabbox}{table}[][\FBwidth]
\usepackage{wrapfig}
\usepackage{algorithm}
\usepackage{algorithmicx}
\usepackage{algpseudocode}

\usepackage{booktabs}
\usepackage{multirow}

\usepackage{enumerate}
\usepackage{amsmath}
\usepackage{multirow}
\usepackage{enumitem}
\usepackage{booktabs}
\usepackage{array}
\usepackage{diagbox}

\usepackage{bm}
\usepackage{verbatim}
\usepackage{svg}
\usepackage{caption}
\usepackage{adjustbox}
\usepackage{subcaption}
\usepackage{caption}
\usepackage[table]{xcolor}

\newcommand{\rebuttal}[1]{\textcolor{black}{#1}}

\title{Scheduling Your LLM Reinforcement Learning with Reasoning Trees}

\author{Antiquus S.~Hippocampus, Natalia Cerebro \& Amelie P. Amygdale \thanks{ Use footnote for providing further information
about author (webpage, alternative address)---\emph{not} for acknowledging
funding agencies.  Funding acknowledgements go at the end of the paper.} \\
Department of Computer Science\\
Cranberry-Lemon University\\
Pittsburgh, PA 15213, USA \\
\texttt{\{hippo,brain,jen\}@cs.cranberry-lemon.edu} \\
\And
Ji Q. Ren \& Yevgeny LeNet \\
Department of Computational Neuroscience \\
University of the Witwatersrand \\
Joburg, South Africa \\
\texttt{\{robot,net\}@wits.ac.za} \\
\AND
Coauthor \\
Affiliation \\
Address \\
\texttt{email}
}

\author{\textbf{Hong Wang}$^{1}$\thanks{Equal Contribution.}\quad
  \textbf{Zhezheng Hao}$^{2}$\footnotemark[1]\quad
  \textbf{Jian Luo}$^{3}$\quad
  \textbf{Chenxing Wei}$^{1}$\\[3pt]
  \textbf{Yao Shu}$^{4}$\quad
  \textbf{Lei Liu}$^{3}$\quad
  \textbf{Qiang Lin}$^{1}$\quad
  \textbf{Hande Dong}$^{1}$\thanks{Corresponding Authors}\quad
  \textbf{Jiawei Chen}$^{2}$\footnotemark[2]\\[3pt]
  $^{1}$ Tencent \quad
  $^{2}$ Zhejiang University \quad
  $^{3}$ Independent Researcher\\
  $^{4}$ Hong Kong University of Science and Technology (Guangzhou)\\
  \texttt{\{hwhongwang, handedong\}@tencent.com}\\
  \texttt{\{haozhezheng, sleepyhunt\}@zju.edu.cn}
}

\newcommand{\ours}{Re-Schedule}

\iclrfinalcopy % Uncomment for camera-ready version, but NOT for submission.
\begin{document}

\maketitle

\begin{abstract}
Using Reinforcement Learning with Verifiable Rewards (RLVR) to optimize Large Language Models (LLMs) can be conceptualized as progressively editing a query's `Reasoning Tree'. This process involves exploring nodes (tokens) and dynamically modifying the model's policy at each node. When combined with data scheduling, this process yields further gains in data efficiency and accuracy.
However, existing RLVR data scheduling methods typically rely on path-based metrics to rank queries, overlooking the reasoning tree structures of these queries.
In this paper, we introduce a novel metric, namely \textbf{Reasoning Score} (r-score), which measures the query's learning difficulty based on the structure of its reasoning tree.
Based on the r-score, we propose the \textbf{\underline{R}easoning Tr\underline{e}e \underline{Schedule}} (\ours), a scheduling algorithm that constructs a curriculum progressing from structurally simple (high r-score) to complex (low r-score) queries.
Experiments on six math-reasoning benchmarks show that {\ours} significantly improves average accuracy, achieving gains of up to 3.2\%.
These strong results validate our approach and demonstrate that a structural understanding of the reasoning tree provides a more powerful and principled foundation for RLVR data scheduling\footnote{Our code is available at https://github.com/zz-haooo/Re-Schedule.}.

\end{abstract}

\section{Introduction}

\begin{figure}[t]
    \centering
    \includegraphics[width=14cm]{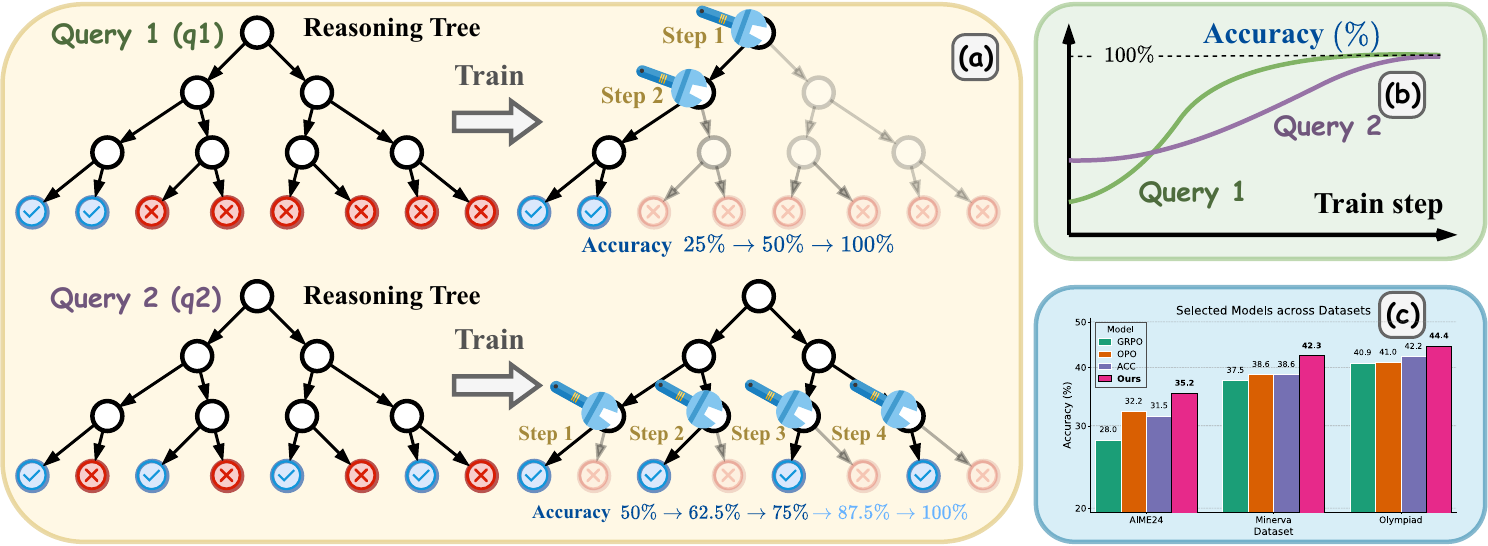}
    \caption{ \textbf{(a)} A simple reasoning tree (q1) requires less node editing for performance improvement than a complex one (q2).
\textbf{(b)} Consequently, q1 shows high training efficiency (steep learning curve) despite low initial accuracy, while q2's complex structure leads to low efficiency.
\textbf{(c)} Our method leverages this structural insight to significantly outperform baselines on various datasets.
}% a)不同query的reasoning tree结构，以及模型在不同qurey上性能的变化,可以看到，Q1的结构单一，只需要少量编辑就可以得到较好的性能，而Q2结构复杂，所以需要多步编辑才能能到较好的性能。 b) LLMs在Q1和Q2上进行RLVR训练过程中的performance变化，对应a)中的树结构，可以看到Q1一开始accuracy较低，但是由于简单的树结构，所以训练效率非常高，虽然Q2有非常多的正确推理路径，然而结构过于复杂，所以需要较长的训练step，因此训练效率低。 c) 在不同数据集上的实验结果验证了我们的这一想法，由于引入推理树的结构作为指标，我们所提的方法显著优于现有工作。注释：为了简化图的结构，我们在这以二叉树的形式呈现，但实际我们探索的并不是二叉树，而是k叉树，理论上来说，若想要得到一个query的k叉树，k应当是词表大小。
    \label{fg_intro1}
\end{figure}

Advancing the complex reasoning capabilities of Large Language Models (LLMs) remains a significant challenge, particularly in domains like mathematical problem-solving. Reinforcement Learning with Verifiable Reward (RLVR)~\citep{1.3_RLVR, 1.1_deepseek_r1}, especially through policy optimization methods like GRPO~\citep{1.6_grpo}, has emerged as a powerful paradigm to address this challenge.
As shown in Figure~\ref{fg_intro1} (a), in this framework, the space of potential solution paths for a query can be modeled as a specific `Reasoning Tree'~\citep{wang2025beyond, yang2025treerpo}, where each node represents an intermediate reasoning step and each path represents a potential solution trajectory. 
From this perspective, RLVR operates as a dynamic `node-editing' process of the reasoning tree: by rewarding correct paths and penalizing incorrect ones, the model iteratively refines its decision policy at each tree node.
This optimization process gradually prunes branches that lead to low-quality or incorrect solutions, thereby improving overall reasoning accuracy.

In this paradigm, data scheduling plays a critical role in model performance~\citep{1.13_data_orz, 1.15_data_limr, yang2026whyattentionpatterns, renuncertainty}.
The concept of data scheduling originates from curriculum learning~\citep{bengio2009curriculum}, 
which posits that models learn more effectively when training examples (queries) are organized in a meaningful sequence. Existing data scheduling strategies typically pre-define a`difficulty' metric for queries, and and schedule them from easy to hard to improve data efficiency and final performance~\citep{r3_add, chen2025data, chen2025seed, dai2025data}
However, from a reasoning tree perspective, current difficulty measure strategies exhibits a critical limitation: current methods estimate difficulty primarily via final solution accuracy, overlooking richer query-level characteristics such as the structural complexity of the reasoning tree. Accuracy alone is insufficient --- low accuracy does not necessarily indicate that a query is inherently hard, and high accuracy does not guarantee ease of optimization. This inconsistency can undermine the efficacy of accuracy-based scheduling approaches. We illustrate this issue with the following examples.

To illustrate, consider two representative queries, \textbf{q1} and \textbf{q2}, whose reasoning trees are shown in Figure~\ref{fg_intro1}(a). As depicted in Figure~\ref{fg_intro1}(b), LLMs may exhibit low initial accuracy on \textbf{q1}, due to the presence of many incorrect solution trajectories (reasoning paths). However, its simple tree structure means that modifying a few key decision nodes can yield substantial accuracy gains, indicating high learning efficiency despite the poor initial performance. In contrast, \textbf{q2} achieves higher initial accuracy, with roughly half of its trajectories being correct, yet these correct paths are scattered across disparate subtrees. This fragmented structure requires more extensive edits across numerous tree nodes, typically resulting in higher training difficulty and lower learning efficiency.
% might yield higher initial accuracy, where half of the reasoning paths are correct. But these correct paths are scattered across disparate sub-trees, resulting in a complex structure that requires more extensive edits for marginal improvement, thus representing low learning efficiency. 
Critically, existing path-based metrics will misinterpret \textbf{q1}'s low accuracy as high difficulty, thus assigning it a lower training weight, while incorrectly prioritizing the more difficult \textbf{q2}. Such path-based metrics may lead to a less efficient training process.
This motivates our central research question: How can we move beyond path-based metrics to directly quantify a query’s true learning difficulty from its reasoning-tree structure?

To address this question, we introduce the \textbf{Reasoning Score} (r-score), a novel metric that quantifies a query's learning potential based on its reasoning tree structure. We formalize this by framing the reinforcement learning training process as an optimization problem under a finite 'node editing budget', which we define as a fixed number of node editing operations. \textbf{Consequently, a query's r-score is its maximum potential accuracy gain achievable within this limited editing budget.} This metric clearly explains the discrepancy in our example: \textbf{q1}, with its `concentrated' error structure, yields a high r-score because a small budget (e.g., two edits) produces a massive accuracy gain (+75\%). Conversely, \textbf{q2}'s `diffuse' structure results in a low r-score, as the same budget only yields marginal improvement (+25\%). Therefore, a higher r-score signifies a more tractable reasoning structure and greater learning efficiency, offering a more comprehensive assessment of difficulty than path-based metrics.

% 基于“可训练性”这一核心概念，我们设计了一种新颖的、基于推理树建模的数据编排算法，以更高效地指导LLM的强化学习过程。该算法主要包括三个阶段：首先，我们通过对基础模型进行多次采样，为每个数据离线构建其推理树的近似表示。其次，我们基于此树状结构，通过模拟剪枝过程计算出每个数据的Trainability。最后，我们将“可训练性”得分作为一个动态权重集成到强化学习的损失函数中。在训练初期，算法会优先为那些Trainability高（即易于学习）的数据分配更高的权重，以帮助模型快速掌握基本模式；随着训练的进行，算法逐渐将重心转移到Trainability较低的数据上，以攻克更复杂的挑战。

% Building on the core concept of reasoning score, we design a novel data scheduling algorithm, \textbf{\underline{R}easoning \underline{C}omplexity \underline{Schedule}} (\ours), based on reasoning tree modeling to guide the LLM's reinforcement learning more effectively. 
% The algorithm comprises three stages: 
% First, we construct an offline approximation of the reasoning tree for each data point by sampling multiple trajectories from the base model. 
% Second, using this tree structure, we calculate each sample's reasoning score by simulating the pruning process. 
% Finally, we integrate reasoning score as a dynamic weight into the RL loss function. 
% During the initial training phase, the algorithm assigns higher weights to high-scoring (easier-to-learn) samples to help the model quickly grasp fundamental patterns. 
% As training progresses, the weights gradually shift toward low-scoring samples to tackle more complex challenges.

Building on the Reasoning Score, we propose the \textbf{\underline{R}easoning Tr\underline{e}e \underline{Schedule} (\ours)}, a novel data scheduling algorithm designed to guide RLVR training more efficiently. Our method consists of three main stages. First, an offline approximation of each query’s reasoning tree is constructed by sampling multiple solution trajectories from a base model. Second, this approximated reasoning tree is used to calculate each query’s reasoning score by simulating the editing process. Finally, we integrate the r-score as a dynamic weight into the RLVR loss function to form a schedule. This schedule prioritizes high-scoring (simple) queries in the initial training phases to accelerate convergence on simple queries. As training progresses, the weighting gradually shifts to lower-scoring (difficult) queries, enabling the model to master more challenging problems.
    
In summary, the main contributions of this paper are:
\begin{itemize}
\item We introduce the Reasoning Score (r-score), a new tree-based metric that measures a query's learning efficiency rather than its path-based solution accuracy.
\item We propose \ours, a data scheduling algorithm that uses the r-score to create an effective, easy-to-hard curriculum for RLVR.
% This offers a new macro-level perspective for understanding the mechanics of RL for LLMs.
\item As shown in Figure~\ref{fg_intro1}(c), we empirically demonstrate that our approach significantly improves average accuracy, achieving gains of up to 3.2\%, on complex reasoning tasks.
\end{itemize}

\section{Related Work}

% RL、GRPO的相关工作；包括 GRPO树状建模的相关工作

% GRPO课程学习的相关工作；包括 GRPO加权、改采样的相关工作

\subsection{Reinforcement learning with verifiable rewards in LLMs}

% Reinforcement Learning with Verifiable Reward (RLVR), where the reward is computed by a rulebased verification function, has been shown to be effective in improving the reasoning capabilities of LLMs. 
% RLVR通常将回答和标准答案对比，得到正确或者错误的二分奖励。
% This reward design obviates the need for complex outcome-based or process-based reward models, offering a straightforward yet potent approach. 
% 近期PPO、GRPO等相关算法的进展进一步优化了这种范式。
% 与这些研究不同，本文在标准GRPO的基础上进行改进，主要聚焦于如何进行更好的raining data schedule。

Reinforcement learning with verifiable reward (RLVR), where the reward is computed by a rule-based verification function, has been shown to be effective in improving the reasoning capabilities of LLMs~\citep{1.3_RLVR,1.1_deepseek_r1,1.2_kimi_k1.5,2.2_rlvr,2.3_rlvr_cur,2.4_rlvr_cur_FastCuRL,wei-etal-2025-flexora, wei-etal-2025-paft,wei2025vflow, wei2025modelgen,maperturbation}. 
Typically, RLVR frameworks assign a binary reward by comparing the model's generated output against a ground-truth solution, indicating whether it is correct or incorrect. This reward design obviates the need for complex outcome-based or process-based reward models, offering a straightforward yet potent approach. Recent advancements in policy optimization algorithms, such as PPO and GRPO, have further refined this paradigm~\citep{1.5_ppo,2.5_2_vineppo,1.19_vc_ppo,2.5_vapo_rlvr,1.6_grpo,1.10_dapo,1.9_UnderstandingR1-Zero-LikeTraining_drgrpo,1.7_SRPO,1.12_REINFORCE++,hao2025rethinking, wei2025redit,liu2026automated,yang2025attentionpredictor}. 
In contrast to these studies, which focus on algorithmic improvements, our work builds upon the standard GRPO framework with a primary focus on designing a more effective training data schedule.

\subsection{Data Scheduling Algorithm in LLM Reinforcement Learning}

Various data scheduling strategies have been proposed to enhance the reasoning capabilities in LLM Reinforcement Learning. These can be broadly categorized into static selection and dynamic adjustment methods.
Representative of static selection is LIMR~\citep{1.15_data_limr}, which selected 1.4k examples from an 8.5k set for RLVR to match the performance of using the full set. 
In contrast, dynamic strategies make real-time adjustments during training. 
For instance, $R^3$ employs reverse curriculum reinforcement learning to simplify the model's exploration space~\citep{r3_add}. 
\rebuttal{LPPO \citep{chen2025data} utilize the gradient of accuracy to prioritize data, effectively treating learning difficulty as a derivative of performance. 
Similarly, Seed-GRPO \citep{chen2025seed} employs semantic diversity (uncertainty) as a proxy for difficulty. 
}
% LPPO dynamically adjusts data weights based on model accuracy and provides online ``prompts" for challenging problems to guide exploration~\citep{chen2025data}. 
% SEED-GRPO focuses on model uncertainty~\citep{chen2025seed}, using token entropy to dynamically re-weight data. 
Furthermore, DELT leverages training gradients to measure the quality and learnability of data~\citep{dai2025data}, subsequently adjusting sample weights.
% CDAS models problem difficulty as a stable trajectory of historical performance to dynamically select problems that align with the model's current competence~\citep{kong2025rethinking}.
% These existing approaches primarily rely on macroscopic, post-hoc metrics derived during the training process, which cannot reveal the intrinsic learning dynamics of a data sample a priori. 
% Our work, in contrast, aims to define a macro metric that can be computed a priori to directly measure the reasoning score (r-score) of each data.
% 现有方法主要依赖于训练过程中得出的宏观事后指标，这些指标无法先验地揭示数据样本的内在学习动态。
% 相比之下，我们的工作旨在定义一个可以先验计算的宏观指标，以直接衡量每个数据的推理得分（R 值）。
\rebuttal{
However, these methods rely on outcome-based proxies (e.g., accuracy), effectively treating reasoning as a flat sequence~\citep{zhao2025redone, zhang2025aligning}. They overlook the inherent tree-structured solution space of reasoning tasks. In contrast, our approach explicitly leverages this topological structure. By analyzing the Reasoning Tree, we directly quantify a query’s `structural learnability', providing a more precise and principled measure of difficulty than performance statistics alone.
}
% However, these existing methods primarily rely on metrics such as accuracy and diversity, which overlook the structural relationships among data samples. 
% In contrast, our work aims to quantify the learning potential of a query by analyzing its reasoning tree.
% 现有方法主要依赖于单路径指标，这些指标没有考虑数据样本之间的结构关系。相比之下，我们的工作旨在从推理树出发量化query的学习潜力

\section{Preliminaries}

% RLVR是什么 GRPO的数学形式
% GRPO的基础流程，loss公式（具体到reward那些东西

% 用树来建模GRPO的相关背景知识 和 一个简单的例子

% This section provides the necessary technical background for our methodology. We first present the mathematical formulation of group relative policy optimization (GRPO) and then describe the 'Reasoning Tree' framework, which is central to our proposed metric.

\subsection{Group Relative Policy Optimization}

The objective of the GRPO algorithm is to optimize a policy $\pi_{\theta}$ based on a group of generated responses~\citep{1.6_grpo, 1.10_dapo}. For a query $q$ from a dataset $\mathcal{D}$, the policy generates $G$ responses $\{o_i\}_{i=1}^{G}$. The token-level objective function is formulated as:
\begin{equation}
\label{eq:grpo_objective}
\mathcal{J}(\theta) = \mathbb{E}_{q \sim \mathcal{D}, \{o_i\}_{i=1}^{G} \sim \pi_{\text{old}}(\cdot | q)} \left[ \frac{1}{\sum_{i=1}^{G} |o_i|} \sum_{i=1}^{G} \sum_{t=1}^{|o_i|} \min \left( r_{i,t} A_{i,t}, \text{clip}(r_{i,t}, 1 - \varepsilon, 1 + \varepsilon) A_{i,t} \right) \right],
\end{equation}
where $r_{i,t} = \frac{\pi_\theta(o_{i,t}|q,o_{i,<t})}{\pi_{\text{old}}(o_{i,t}|q,o_{i,<t})}$ is the probability ratio of the token $o_{i,t}$ between the current and old policies. The advantage term $A_{i,t}$ is constant for all tokens within a single response and is calculated by normalizing the response's reward $R_i$ relative to the other responses in the group:
\begin{equation}
\label{eq:advantage}
A_{i,t} = \frac{R_i - \text{mean}(\{R_k\}_{k=1}^G)}{\text{std}(\{R_k\}_{k=1}^G) + \delta}, \quad \forall t,
\end{equation}
where $\delta$ is a small constant for numerical stability. 
% To prevent the policy from deviating too far from a trusted reference policy $\pi_{\text{ref}}$, a KL-divergence penalty is added to the objective.

Data scheduling algorithms can be formulated by introducing a weighting function $\omega(q,t)$ that modulates the contribution of each query $q \in \mathcal{D}$ and current epoch $t$ to the overall objective. 
Specifically, the objective in Equation~\ref{eq:grpo_objective} is modified as follows:
\begin{equation}
\label{eq:scheduled_objective_alt}
\mathcal{J}_{\text{schedule}}(\theta) = \mathbb{E}_{q \sim \mathcal{D}, \dots} \left[ \omega(q,t) \cdot \left( \text{original objective term for } q \right) \right].
\end{equation}
Note: In the equations above, we have abbreviated the full objective for clarity. 
% 例如 基于正确率的课程学习训练权重 $\alpha$ 被表述为
% \begin{equation}
% \alpha(ACC(q), t) = (1 - \gamma(t)) ACC(q) + \gamma(t) (1-ACC(q))
% \end{equation}
% 常见的$\gamma(t)$ 可以设计为线性映射 $\gamma(t) = \frac{t}{T}$ 或 S 型函数 $\gamma(t) = \sigma \left( k \left( \frac{t}{T} - 0.5 \right) \right)$
For example, in an accuracy-based curriculum learning, the training weight $\omega$ is formulated as a function of the query's accuracy $\text{ACC}(q)$ and current epoch $t$:
\begin{equation}
% {\scriptsize
\alpha(\text{ACC}(q), t) = (1 - \gamma(t)) \text{ACC}(q) + \gamma(t) (1-\text{ACC}(q)), 
\end{equation}
\begin{equation}
\ \omega = \text{rank}(\alpha)\% \cdot \omega_{\text{max}} + (1-\text{rank}(\alpha)\%) \cdot \omega_{\text{min}}.
% }
\end{equation}
Here, $\omega_{\text{max}}$ and $\omega_{\text{min}}$ are hyperparameters that define the maximum and minimum training weights (e.g., $\omega_{\text{max}}=0.8$, $\omega_{\text{min}}=0.2$); 
And $\text{rank}(\alpha)$ means calculating the reverse order of $\alpha$ in the entire dataset.
The term $\gamma(t)$ is a scheduling function that progresses over time. 
Common choices for $\gamma(t)$ include a linear mapping, $\gamma(t) = t/T$, or a sigmoid function, $\gamma(t) = \sigma \left( \left( \frac{t}{T} - 0.5 \right) \right)$, $ \sigma(x) = (1+e^{-x})^{-1}$, where $T$ is the total number of epochs.

% You should insert the full term from Equation (1) inside the expectation. For example, you can write it as:
% \mathcal{J}_{\text{schedule}}(\theta) = \mathbb{E}_{q \sim \mathcal{D}, \{o_i\}_{i=1}^{G} \sim \pi_{\text{old}}(\cdot | q)} \left[ \alpha(q) \cdot \left( \frac{1}{\sum_{i=1}^{G} |o_i|} \sum_{i=1}^{G} \sum_{t=1}^{|o_i|} \min \left( r_{i,t} A_{i,t}, \text{clip}(r_{i,t}, 1 - \varepsilon, 1 + \varepsilon) A_{i,t} \right) \right) \right]

% 现有Data Scheduling Algorithm可以视为不同数据$q \in \mathcal{D}$的目标函数乘上一个权重函数\alpha(q)。即 \mathcal{J}(\theta)_{Schedule} = \mathbb{E}_{q \sim \mathcal{D}, \{o_i\}_{i=1}^{G} \sim \pi_{\text{old}}(\cdot | q)} \alpha(q) \left[ \cdot \right].

% \begin{figure}[t]
%   \centering
% \hspace{-0.5cm}
%   % 第一张图片
%   \begin{minipage}{0.5\textwidth}
%     \centering
%     \includegraphics[width=1\linewidth]{figure/average_trajectories_comparison.pdf}
%   \end{minipage}
%   % 第二张图片
%   \begin{minipage}{0.5\textwidth}
%     \centering
%     \includegraphics[width=1\linewidth]{figure/intro2_5.pdf}
%   \end{minipage}
%   \caption{\textbf{Left.} 
%   .
%   \textbf{Right.} 
%   .
%   }
%   \label{fg_motivation1}
% \end{figure}
% % \vspace{-0.5cm}

\subsection{Reasoning Tree}

% For complex reasoning tasks, the process of generating a solution can be conceptualized as traversing a 'Reasoning Tree'. In this abstraction, the root of the tree is the initial prompt, and each node represents a partial solution or an intermediate reasoning step. The branches extending from a node correspond to the possible next tokens or thought segments that the LLM can generate. A complete path from the root to a leaf node forms a full reasoning trace and culminates in a final answer.

% Some paths in this tree lead to correct solutions (and thus receive a positive reward), while many others lead to incorrect outcomes. 
% The goal of the RLVR training process, from this perspective, is to optimize fork, i.e., prune erroneous paths.
% By rewarding correct paths and penalizing incorrect ones, the policy optimization algorithm adjusts the token probabilities at each node, effectively strengthening the branches that lead to correct answers and weakening those that lead to errors. The structure of this tree---the distribution of correct and incorrect paths---is intrinsic to each problem sample and, as we will argue, is a key determinant of its learning dynamics.

For complex reasoning tasks, the process of generating a solution can be conceptualized as traversing a `Reasoning Tree'. In this context, the root of the tree is the initial prompt, and each node represents a partial solution or an intermediate reasoning step. The branches extending from a node correspond to the possible next tokens or thought segments that the LLM can generate. 

Due to the combinatorial explosion of possible solution paths, the complete reasoning tree is typically computationally intractable. Therefore, analysis often relies on a structured approximation (e.g., a fixed-structure k-ary reasoning tree). Formally, an approximated reasoning tree is defined as a triplet $T = (\mathcal{N}, \mathcal{E}, \mathcal{R})$, where $\mathcal{N}$ is the set of nodes, $\mathcal{E}$ is the set of edges, and $\mathcal{R}$ defines the parent-child relationships.

\begin{wrapfigure}{r}{0.4\textwidth} % r 表示图片在右侧，0.5\textwidth 表示图片占页面宽度的一半
% \vskip -0.10in
    \centering
    \includegraphics[width=\linewidth]{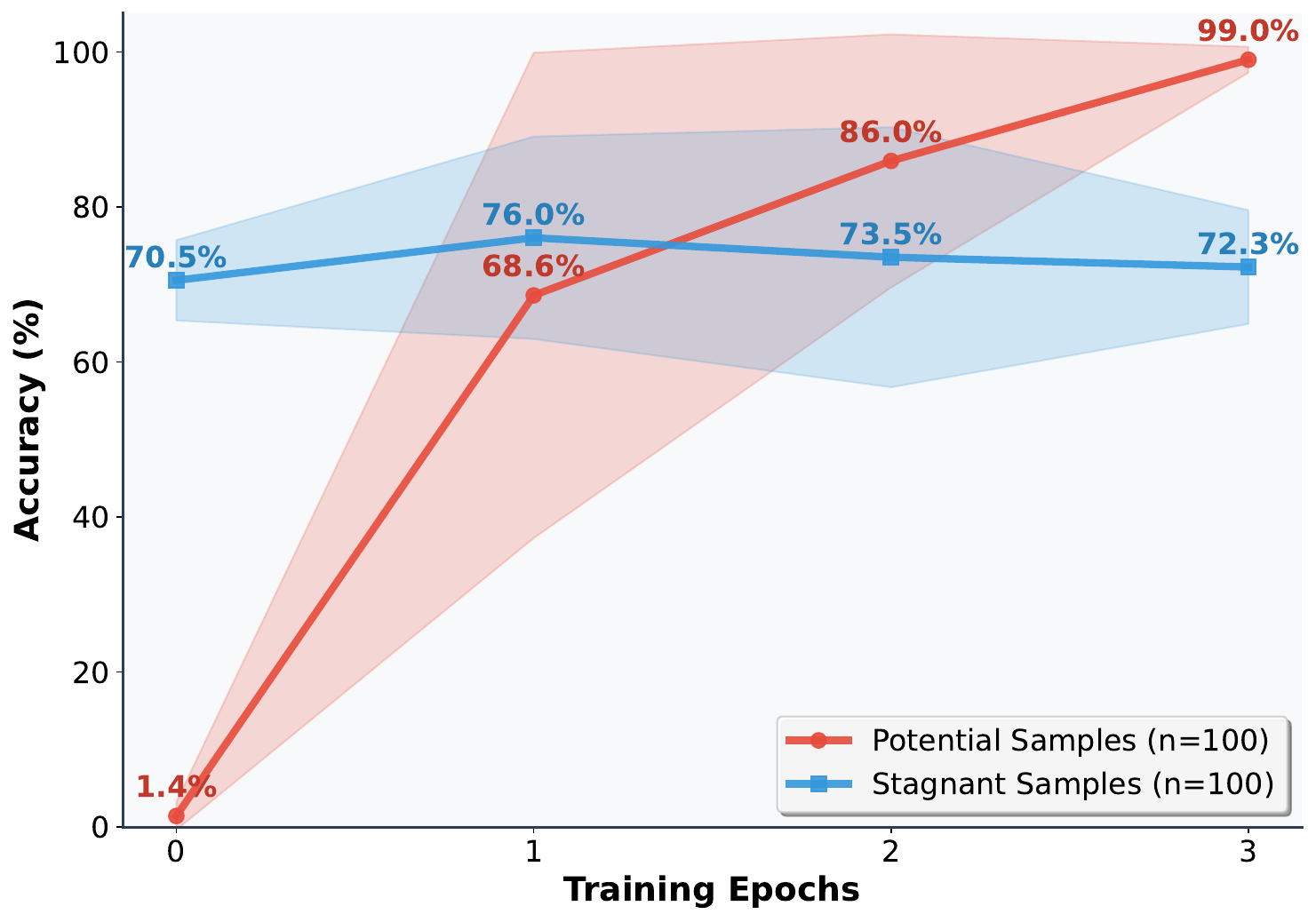}
    \caption{Accuracy Progression During Training. The solid line represents the average accuracy, and the shaded area indicates the range.} 
    % \rebuttal{The red line (Potential Samples) corresponds to high r-score queries (structurally simple), while the blue line (Stagnant Samples) corresponds to low r-score queries (structurally complex)} 
    % The experiment was conducted using the Qwen2.5-Math-7B model on the DAPA-Math-17K dataset.
    % 部分数据在训练过程中的正确率变化曲线，其中实线表示这些数据的平均正确率，阴影表示正确率的范围。本实验在qwen 2.5 math 7B，DAPA-Math-17K 数据集上进行。
    \label{fg_motivation1}
    \vskip -0.10in

\end{wrapfigure}

The components of the tree are described using the following notation: $\mathcal{N}_{\text{leaf}} \subset \mathcal{N}$ is the set of leaf nodes;
For a given node $n_i \in \mathcal{N}$, $C(n_i)$ denotes the set of its immediate children and $\mathcal{L}(n_i)$ denotes the set of its leaf descendants.
If $n_i$ is a leaf node, then $\mathcal{L}(n_i) = \{n_i\}$.
Within this framework, each non-leaf node $n_i \in \mathcal{N} \setminus \mathcal{N}_{\text{leaf}}$ represents a partial reasoning path, while a complete path to a leaf node $n_j \in \mathcal{N}_{\text{leaf}}$ corresponds to a full solution trajectory.

% Some paths in this tree lead to correct solutions (and thus receive a positive reward), while many others lead to incorrect outcomes. 
From this perspective, the RLVR optimization process is a dynamic `node editing' of this reasoning tree. 
By rewarding correct paths and penalizing incorrect ones, the policy optimization algorithm adjusts the token probabilities at each node, effectively strengthening the branches that lead to correct answers and weakening those that lead to errors. 
The structure of this tree---the distribution of correct and incorrect paths---is intrinsic to each problem sample and, as we will argue, is a key clue to its learning dynamics.

\section{Motivation}

% fg_motivation1

% 本文的出发点是发现用正确率等这种指标直接衡量一个问题的训练难度是不合适的。
% 作为intro中举例的补充，如图2所示，我们在整个数据集中各自筛选了100个数据。其中红色的表示在训练初期有很高的正确率，但经过较长步数的训练才提升正确率的例子。而蓝色的反之。
% 虽然，前期红色的数据正确率一直很高，但是蓝色的数据代表学习难度更低。这说明了正确率并非一个良好的评价数据学习难度的指标。这也启发了我们从推理树的角度设计指标。

The premise of this work is that path-based metrics such as accuracy are poor indicators of a query's true learning difficulty. To illustrate our point, we supplement the example from the introduction with an experiment. 
As shown in Figure \ref{fg_motivation1}, we selected two distinct sets of 100 queries each from the DAPA-Math-17K dataset, using the Qwen2.5-Math-7B model.
\rebuttal{
The blue line represents `Stagnant Samples'—queries with high initial accuracy but complex reasoning structures (low r-score). Their flat learning curve indicates that despite high initial performance, they are difficult to improve further. In contrast, the red line represents `Potential Samples'—queries with low initial accuracy but simple tree structures (high r-score). Their steep learning curve demonstrates high learnability, where a small amount of training yields significant gains.
}
% The blue line represents queries that have high initial accuracy but require extensive training to show further improvement. 
% In contrast, the red line, standing for queries with simple tree structures, begins with low accuracy but improves rapidly. 
% Although queries for the blue line appear `easier' based on their initial accuracy, queries for the red line demonstrate greater potential. 
This discrepancy highlights that path-based metrics, like accuracy, are biased measurements for learning difficulty. This finding motivates us to design a new metric based on the structure of the reasoning tree.

% 为了进一步说明这一点，我们进行了以下分析

% We define a query's the ease of training over a given training interval as the magnitude of its accuracy improvement. 
% Queries with the largest accuracy gains are thus considered the most learnable within that period. 

% we compare several metrics—our r-score, initial accuracy—on their ability to identify the most learnable queries. 
% At a specific training step, we use each metric to select the top 10\% of queries predicted to be the most learnable. 
% We then measure the intersection of these selected sets with the ground-truth set of most learnable queries (i.e., those that actually had the highest accuracy gains).

% As depicted in Figure 4, the set selected via r-score has the largest intersection with the ground-truth set. This result indicates that r-score is a more accurate predictor of a query's true learnability than the other metrics.

% 这也启发了我们的算法设计

\section{Method}

\begin{figure}[t]
    \centering
    \includegraphics[width=14cm]{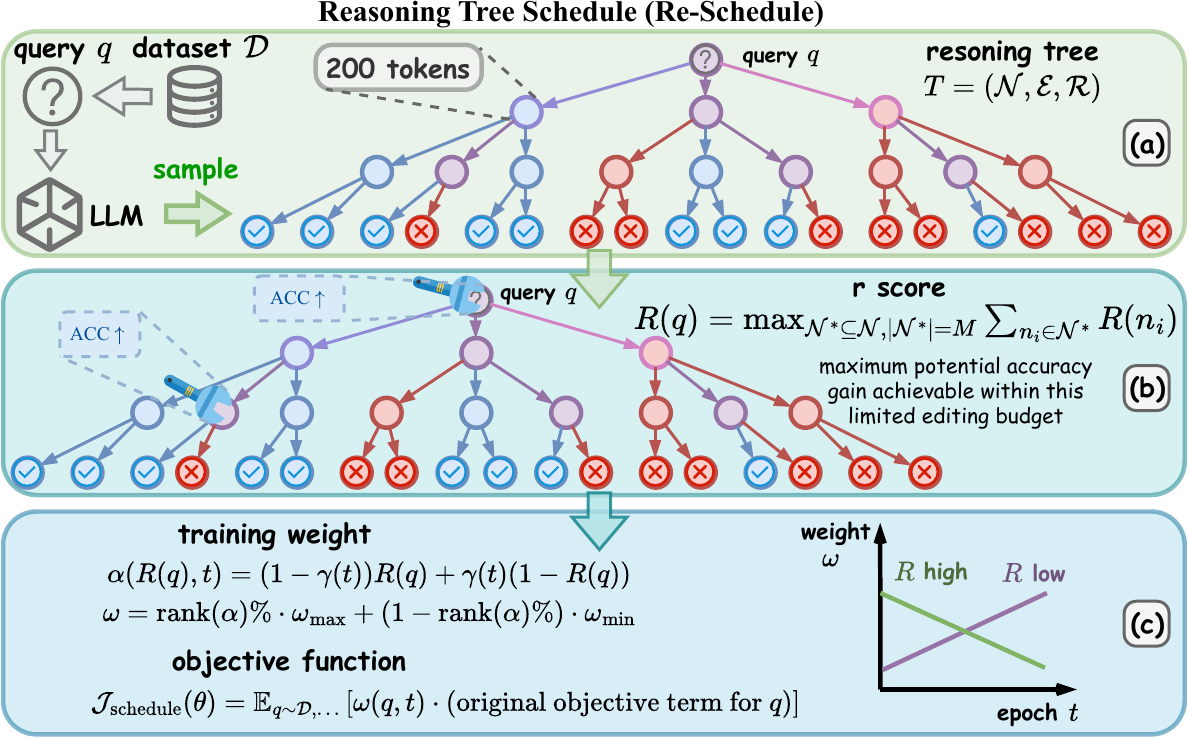}
    \caption{Overview of the \textbf{Reasoning Tree Schedule (Re-Schedule)} Algorithm.\textbf{(a) Tree Construction}: For each query, an approximate reasoning tree is constructed by sampling multiple solution paths from a base model \rebuttal{(Note: This figureis for illustrative purposes only; our experiments use a tree with a depth of 4 and a width of 4, i.e., $k=4, d=4$.)}.
\textbf{(b) R-Score Calculation}: The tree's structure is analyzed to compute the r-score, a metric quantifying the query's learning potential.
\textbf{(c) Dynamic Weighting}: The r-scores are used to dynamically weight each query during training, forming a curriculum that progresses from structurally simple (easy) to complex (hard) examples.}
    % Reasoning Tree Schedule (Re-Schedule) 算法流程图：（a）our approach begins by constructing an approximate reasoning tree for each query by sampling multiple solution paths from a base model. (b) This tree's structure is then analyzed to compute the r-score, which measures the query's learning potential. (c) Finally, these r-scores are used to dynamically weight each query during training, creating a curriculum that progresses from structurally simple (easy) to complex (hard) examples.
    % （a）我们对每个query进行采样，得到一棵推理树（b）通过树结构，计算该query的r score（c）训练时，利用query的r score对其进行加权
    \label{fg_method1}
\end{figure}

% 总起
% 树如何构建
% 如何计算指标
% 如何进行加权

% 在本文中，我们从推理树的结构出发，提出了一个能够先验地度量数据学习难度的指标——Reasoning Score (r-score)。基于此，我们设计了一种提升强化学习性能的数据编排算法， Reasoning Complexity Schedule (RC-Schedule) 。如图2所示，我们的算法主要分为三个步骤：
% 1. 构建推理树：用待训练的base模型，对每个query构建近似的推理树，并判别每个回答对错
% 2. 计算r-score：通过每个数据的推理树，计算每个数据的r-score
% 3. r-score加权：在训练的过程中，动态给每个数据计算权重。实现训练初期，重点关注 r-score高的题目。训练后期，重点关注 r-score低的题目。
% 下面我们先详细介绍这三个主要步骤，然后分析 r-score指标的合理性
% In this paper, we introduce a novel metric, the Reasoning Score (r-score), which is derived from the structure of the reasoning tree to a priori quantify the learning difficulty of a data sample. Building on this, we design a data scheduling algorithm to enhance reinforcement learning performance, named the Reasoning Tree Schedule (Re-Schedule). 
% As illustrated in Figure~\ref{fg_method1}, our algorithm comprises three main steps:
% \begin{itemize}
%     \item [1.] Reasoning Tree Construction: For each query, we use the base model to construct an approximate reasoning tree and verify the correctness of each generated response.
%     \item [2.] r-score Calculation: We compute the r-score for each query based on the structure of its corresponding reasoning tree.
%     \item [3.] r-score Weighting: During training, we dynamically assign a weight to each sample. This curriculum prioritizes samples with high r-scores in the early stages and shifts focus to those with low r-scores in the later stages.
% \end{itemize}

As illustrated in Figure~\ref{fg_method1}, the \textbf{Reasoning Tree Schedule (Re-Schedule)} enhances reinforcement learning performance by creating a curriculum based on our novel metric, the \textbf{Reasoning Score (r-score)}. 
The r-score quantifies a query's learning difficulty {a priori} based on the structure of its reasoning tree. 
Next, we will introduce the specific implementation details.
% As illustrated in Figure~\ref{fg_method1}, our approach begins by constructing an approximate reasoning tree for each query by sampling multiple solution paths from a base model. This tree's structure is then analyzed to compute the r-score, which measures the query's learning potential. Finally, these r-scores are used to dynamically weight each query during training, creating a curriculum that progresses from structurally simple (easy) to complex (hard) examples.

% As illustrated in Figure~\ref{fg_method1}, our approach consists of three main stages:
% \begin{enumerate}
%     \item \textbf{Reasoning Tree Construction:} For each query, we use a base model to construct an approximate reasoning tree by sampling multiple solution paths.
%     \item \textbf{r-score Calculation:} We compute an r-score for each query, which measures its learning potential based on the tree's structure.
%     \item \textbf{r-score Weighting:} We use the r-score to dynamically weight each sample during training, creating a curriculum that progresses from easy to hard examples.
% \end{enumerate}

\subsection{Tree Construction}

As the entire reasoning tree is computationally intractable, we construct a manageable, fixed-structure $k$-ary approximation for each query $q$. The structure of this tree, $\mathcal{T}$, is defined by a branching factor $k$, a maximum depth $d$, and a token interval $l$ (e.g., $k=4, d=4, l=200$).

The construction process begins at the root node (the query $q$) and proceeds via a periodic branching strategy during response generation. Specifically, a branch is triggered immediately at the beginning of the response and subsequently at every $l$-token interval. As shown in Figure \ref{fg_method1} (a), at each trigger, the current path splits into $k$ independent sub-paths that continue to generate in parallel. This recursive branching process continues until a predefined maximum depth $d$ is reached. To minimize computational overhead from this multi-path sampling, we use the Key-Value (KV) Cache, as all sibling branches share the same prefix.

In RLVR tasks, a solution's quality is determined by the correctness of its final answer, which corresponds to a leaf node in our framework. Therefore, we define the quality of any intermediate node $n_i$ as the average accuracy of its leaf descendants, $\mathcal{L}(n_i)$. This is quantified using an accuracy function:
\begin{equation}
\label{eq:acc_func}
\text{ACC}(S) = \frac{\sum_{n_j \in S} \mathbb{I}(n_j \text{ is correct})}{|S|},
\end{equation}
where $S$ is a set of leaf nodes and $\mathbb{I}(\cdot)$ is the indicator function. This allows us to assess quality at different levels: the quality of a reasoning segment via $\text{ACC}(\mathcal{L}(n_i))$ and the model's aggregate performance on the query via $\text{ACC}(\mathcal{N}_{\text{leaf}})$.

\subsection{R-Score Calculation}

% Optimizing a node on a reason tree is equivalent to choose a certain child for this node.
% Further, how easy a node can be trained during reinforcement learning can be evaluated by: the maximal accuracy improvement of the whole reasoning tree by choosing one of its child.
% Therefore, given a non-leaf node $n_i$, its Reasoning Score $R(n_i)$ is defined as
% \begin{equation}
%     R(n_i) = \max_{n_\text{child} \in \mathcal{C}(n_i)} \text{ACC} \big[\mathcal{N}_\text{leaf} \setminus \mathcal{L}(n_i) \cup \mathcal{L}(n_\text{child})\big] - \text{ACC}\big[\mathcal{N}_\text{leaf}\big].
% \end{equation}
% The overall Reasoning Score for query $q$ is 
% \begin{equation}
%     R(q) = \max_{\mathcal{N}^{*} \subseteq \mathcal{N}, |\mathcal{N}^{*}|=M} \sum_{n_i \in \mathcal{N}^{*}}  R(n_i), \text{ where } \mathcal{N} \text{ is generated by }q.
% \end{equation}
% Intuitively, the metric $R(q)$ measures how much accuracy that LLM can be improved on query $q$ with limited optimization during RL training.
% $M$ is a hyperparameter to control the degree of training, e.g., $M=4$.

The r-score quantifies the learning potential of a node or query by measuring the maximum achievable accuracy gain under a limited policy refining cost, like a limited node editing budget.
% For a non-leaf node $n_i$, this corresponds to selecting the child branch that yields the highest overall tree accuracy. We define its r-score, $R(n_i)$, as this maximal accuracy improvement:
% \begin{equation}
%     R(n_i) = \max_{n_\text{child} \in \mathcal{C}(n_i)} \text{ACC} \big[\mathcal{N}_\text{leaf} \setminus \mathcal{L}(n_i) \cup \mathcal{L}(n_\text{child})\big] - \text{ACC}\big[\mathcal{N}_\text{leaf}\big].
% \end{equation}
% The overall r-score for a query, $R(q)$, estimates the total accuracy gain achievable under a limited optimization budget, represented by correcting a maximum of $M$ nodes. It is the maximum sum of node r-scores over any subset of $M$ nodes (e.g., $M=4$):
% \begin{equation}
%     R(q) = \max_{\mathcal{N}^{*} \subseteq \mathcal{N}, |\mathcal{N}^{*}|=M} \sum_{n_i \in \mathcal{N}^{*}}  R(n_i).
% \end{equation}
% % 注：任意n_i1，n_i2节点\in N，n_i1\neq n_i2 且n_i1不在R（n_i2）选择过程中不被选择的分支上
Given this idea, for any non-leaf node $n_i$, we define its r-score, $R(n_i)$, as the maximal accuracy gain achievable by selecting its single best child branch and pruning all others. This is formulated as:
\begin{equation}
\label{eq:r-score}
    R(n_i) = \max_{n_\text{child} \in \mathcal{C}(n_i)} \text{ACC} \big[\mathcal{N}_\text{leaf} \setminus \mathcal{L}(n_i) \cup \mathcal{L}(n_\text{child})\big] - \text{ACC}\big[\mathcal{N}_\text{leaf}\big].
\end{equation}
The overall r-score for a query, $R(q)$, estimates the total accuracy gain achievable under a budget that limits modifications to a maximum of $M$ nodes. It is the maximum sum of r-scores from any set of $M$ non-conflicting nodes (e.g., for a budget of $M=4$):
\begin{equation}
\label{eq:r-score-sum}
    R(q) = \max_{\mathcal{N}^{*} \subseteq \mathcal{N}, |\mathcal{N}^{*}|=M} \sum_{n_i \in \mathcal{N}^{*}} R(n_i).
\end{equation}
Two nodes are considered conflicting if one is located in a subtree that is implicitly pruned by the optimal branch selection of the other. 

Intuitively, solving Equation (\ref{eq:r-score}) represents the evaluation process of the sub-tree's structure, while a simpler structure of reasoning tree starting from $n_i$ yields a higher $R(n_i)$. Combining the evaluation $R(n_i)$ of each node $n_i$ under a limited budget $M$, solving Equation (\ref{eq:r-score-sum}) is to find the maximum achievable accuracy gain over the reasoning tree, like exploring possible combinations of $M$ nodes and picking the best combination. Thus, a higher $R(q)$ indicates that substantial accuracy improvements can be made by correcting just a few critical reasoning steps, signifying a structurally simple and efficient-to-learn query.

\subsection{Dynamic Weighting}

To strike a balance between data diversity and data scheduling, we propose a weighted scheduling framework that dynamically adjusts data prioritization.
Specifically, queries are assigned adaptive weights determined by both training step $t$ and r-score $R$. Specifically, when it is an early training stage, higher weights are assigned to samples with higher r-scores (indicating lower learning difficulty), stabilizing the reinforcement learning. When RL training meets the later training phase, queries' weights will be redistributed gradually towards lower-r-score samples (higher learning difficulty) to enhance model generalization.

Motivated by this, the training weight $\omega$ of each query is formulated as
% \begin{equation}
% \alpha(R(q), t) = (1 - \gamma(t)) R(q) + \gamma(t) (1-R(q)), \quad \omega = \alpha \cdot \omega_{\text{max}} + (1-\alpha) \cdot \omega_{\text{min}},
% \label{eq:w(q)}
% \end{equation}
\begin{equation}
% {\scriptsize
\alpha(R(q), t) = (1 - \gamma(t)) R(q) + \gamma(t) (1-R(q)), 
\end{equation}
\begin{equation}
\omega = \text{rank}(\alpha)\% \cdot \omega_{\text{max}} + (1-\text{rank}(\alpha)\%) \cdot \omega_{\text{min}},
% }
\label{eq:w(q)}
\end{equation}
where $t$ is the current epoch; $\omega_{\text{max}}$ and $\omega_{\text{min}}$ are hyperparameters that define the maximum and minimum training weights;
And $\text{rank}(\alpha)$ means calculating the reverse order of $\alpha$ in the entire dataset;
$\gamma(t)$ can be either linear mapping $\gamma(t) = \frac{t}{T}$ or sigmoid $\gamma(t) = \sigma \left( \left( \frac{t}{T} - 0.5 \right) \right)$.
The $\alpha(R(q), t)$ is a monotonically varying function that down‐weights high-scoring (simple) queries over time while up‐weighting lower-scoring (difficult) ones.
This scheduling approach balances exploitation of easily learnable patterns and exploration of challenging instances, mitigating catastrophic forgetting of underrepresented data distributions.

\section{Analysis}

% 为了进一步说明我们算法的合理性，我们需要验证设计过程中的两个假设：
% 1. 训练过程是推理树的优化分支的过程。
% 2. Reasoning Score和query的训练容易程度正相关。
% 下面我们依次验证这两点

% To substantiate the rationale behind our proposed method, we validate two core assumptions that underpin its design:
% \begin{itemize}
%     \item The RL training process functions as an optimization of the reasoning tree, progressively optimizing forks.
%     \item Reasoning Score (r-score) is positively correlated with learnability.
% \end{itemize}

% We now proceed to validate each of these assumptions empirically.

% To further demonstrate the rationale for r-score, this section conducts two analytical experiments.

\subsection{Training as Reasoning Tree Optimization}

To empirically validate that the training process is optimizing reasoning trees, we conducted an experiment centered on a new metric: the Minimum Corrective Nodes (MCN). 
This metric is defined as the minimum number of node modifications required for the reasoning tree to achieve a specified target accuracy. 
A single node modification is counted as one token change; thus, a lower MCN signifies a well-structured reasoning tree. 
In our experiment, we tracked the MCN on the DAPA-Math-17K training set during the training of Qwen2.5-Math-7B, excluding queries where the base model's accuracy was below 10\%.

% 如图 3 （a）所示，随着训练的进行，不论指定正确率是多少，训练集上的平均 MCN 呈现持续下降趋势。这一结果表明强化学习过程成功地改进了模型在关键决策点的策略，从而验证了我们的核心论点：训练是推理树优化的过程。

\begin{figure}[t]
    \centering
    \includegraphics[width=14cm]{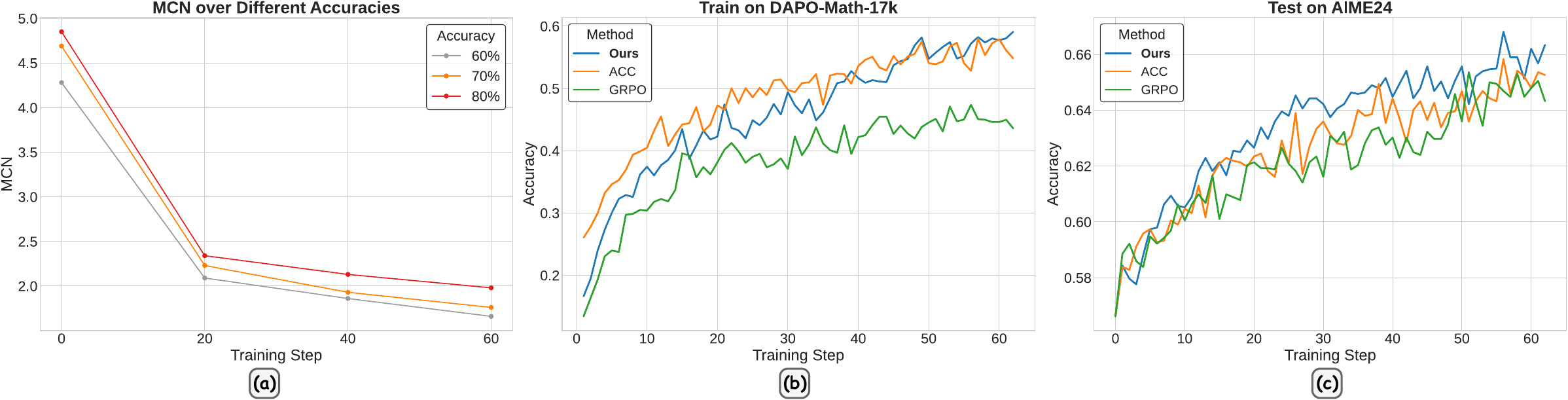}
    \caption{\textbf{(a)} The average MCN decreases over time, indicating successful tree optimization.
\textbf{(b)} \& \textbf{(c) }To compare metrics, we train models on the top 1/3 of data selected by each. The plots show the resulting \textbf{(b)} training accuracy and \textbf{(c)} test accuracy. The model used is Qwen2.5-Math-7B.}
    % (a) 训练过程中，平均 Minimum Corrective Nodes (MCN)的变化曲线； 筛选不同指标下最容易学习的1/3个数据进行训练，（b）在训练集上的平均正确率变化曲线，（C）在测试集上的平均正确率变化曲线。模型选用qwen 2.5 math 7B
    \label{fg_an1}
\end{figure}

As shown in Figure \ref{fg_an1}(a), the average MCN across the training set exhibits a consistent downward trend as training progresses, regardless of the target accuracy. This result demonstrates that the reinforcement learning process effectively refines the model's policy at critical decision nodes, thereby validating our central assumption that training is a process of reasoning tree optimization.

\subsection{The relationship between R-Score and Learning Difficulty}

In this experiment, we want to see which metric best identifies valuable queries for early-stage training. The process is as follows: First, we use each metric to select the top one-third of the data, creating several distinct subsets. Second, we train a separate model on each of these subsets for a single epoch. Finally, we evaluate the resulting models on both the training and test sets.

As shown in Figure \ref{fg_an1}(b), the subset selected by the ACC-based method initially shows higher average accuracy on the training set, as expected from its selection criteria. 
However, as training progresses, the model trained on the r-score-selected subset quickly surpasses it. This indicates that the r-score is more effective at identifying queries with low learning difficulty, rather than just initial accuracy.

The advantage of r-score is even more evident on the test set, as shown in Figure \ref{fg_an1}(c). 
Here, the model trained on the r-score-selected queries consistently outperforms both the ACC-based selection and a baseline with random query selection (GRPO). 
This confirms that the queries identified by the r-score provide the most effective learning signal, leading to better performance improvement and validating its capability in identifying the real difficulty of queries.

\section{Experiment}

% 为了全面说明，我们提出算法的性能，我们和相关工作进行了全面的对比。
% 

\subsection{RL training Setups}

\textbf{Training setting} We conduct experiments on two different models, including Qwen2.5-Math-7B and Qwen2.5-7B.
We adapt our training codebase from verl \citep{sheng2025hybridflow} and follow the training recipe of standard GRPO.
Our training data is DAPO-Math-17k \citep{1.10_dapo}, containing only math problems with integer ground-truth answers.
Both the KL-divergence and entropy loss terms are removed in our experiments~\citep{hao2025rethinking}.
Generation batch size is set to $512$. 
Training is performed with top-p value of $1.0$ and temperature $=1.0$.
% Detailed training hyperparameters of our method and baselines are in Appendix \ref{training details}.
% 

\textbf{Evulation}
We evaluated our models and baselines on six widely used mathematical reasoning benchmarks: AIME24, AIME25, AMC23 \citep{4_11_Numinamath}, MATH-500 \citep{4_3_math_dataset}, Minerva Math \citep{4_4_Minerva_dataset}, and OlympiadBench \citep{4_5_OlympiadBench}.
Validation is performed with a top-p value of $0.7$ and temperature $=1.0$ across all models and test sets.
We use Math-Verify for training, validation, and final evaluation.
\rebuttal{We report avg@32 for all datasets. All results are presented as percentages.}

\textbf{Baselines}
For the throughout comparison, we compare our method against 7 baselines, including standard GRPO \citep{1.6_grpo}, SimpleRL-Zoo \citep{2.2_rlvr}, Eurus-PRIME\citep{4_10_Eurus_prime}, OPO \citep{hao2025policy}, 
ACC (curriculum learning based on accuracy, using sigmoid weighting), LPPO~\citep{chen2025data}, and Seed-GRPO~\citep{chen2025seed}.

\textbf{Our Methods}
% 我们的算法Re-Schedule采用linear和sigmoid两种加权方式，以下实验中除非特殊说明，Re-Schedule的树采用 branching factor $k=4$, a maximum depth $d=4$, and a token interval $l=200$ 的参数。
% 其中 ‘linear’ 采用 $\gamma(t) = \frac{t}{T}$的加权方式；‘sigmoid’采用 $\gamma(t) = \sigma \left( \beta \left( \frac{t}{T} - 0.5 \right) \right)$的加权方式.
% 其中T设置为10，\beta为1。
Re-Schedule is implemented with two weighting schemes: `linear' and `sigmoid'. Unless otherwise specified, the reasoning trees in our experiments are constructed with a branching factor of $k=4$, a maximum depth of $d=4$, and a token interval of $l=200$. The weighting schemes are defined as follows:
1. The `linear' scheme uses $\gamma(t) = t/T$;
2. The `sigmoid' scheme uses $\gamma(t) = \sigma \left(\left( \frac{t}{T} - 0.5 \right) \right)$.
For both, we set the total number of epochs $T=10$.
Details of the training setup can be found in the Appendix \ref{exp_main}.

\subsection{Main Experiment}

\begin{table*}[h]
\centering
{\setlength{\tabcolsep}{3pt}
\renewcommand{\arraystretch}{1.2}
\begin{tabular}{lccccccc}
\hline
\textbf{Model} & \textbf{AIME24} & \textbf{AIME25} & \textbf{AMC23} & \textbf{MATH500} & \textbf{Minerva} & \textbf{Olympiad} & \textbf{Avg.} \\
\hline
Qwen2.5-Math-7B & 13.8 & 5.3 & 44.6 & 39.6 & 9.9 & 13.8 & 21.2 \\
\hline
\multicolumn{8}{c}{\textbf{Classical RLVR Methods}} \\
\hline
GRPO            & 28.0 & 14.3 & 66.2 & 78.6 & 37.5 & 40.9 & 44.3 \\
\rebuttal{SimpleRL-Zoo}    & \rebuttal{30.8} & \rebuttal{14.2} & \rebuttal{65.4} & \rebuttal{79.2} & \rebuttal{37.1} & \rebuttal{40.8} & \rebuttal{44.6} \\
Eurus-PRIME     & 20.9 & 13.0 & 65.2 & 79.8 & 37.5 & 40.6 & 42.8 \\
OPO             & 32.2 & 13.4 & 71.5 & \textbf{82.2} & 38.6 & 41.0 & 46.5 \\
\hline
\multicolumn{8}{c}{\textbf{Scheduling Methods}} \\
\hline
$\text{ACC}_{sigmoid}$              & 31.5 & 15.6 & 70.9 & 80.8 & 38.6 & 42.2 & 46.6 \\
LPPO       & 32.8 & 14.9 & 63.3 & 79.2 & 39.0 & 40.6  & 45.0  \\
Seed-GRPO       & 30.7 & 14.0 & 71.0 & 80.0 & 38.2 & 38.5 & 45.4 \\
\hline
\multicolumn{8}{c}{\textbf{Our Methods}} \\
\hline
% linear\_min\_0.5\_max\_2.0\_T\_10   & 34.2 & 15.6 & 72.4 & 81.2 & 36.4 & 42.5 & 47.1 \\
% sigmoid\_min\_0.5\_max\_2.0\_T\_10  & 35.2 & 16.0 & 72.3 & 82.2 & 42.3 & 44.4 & 48.5 \\
\rowcolor{gray!10} $\text{Re-Schedule}_{linear}$   & 34.2 & 15.6 & \textbf{72.4} & 81.2 & 36.4 & 42.5 & 47.1 \\
\rowcolor{gray!10} $\text{Re-Schedule}_{sigmoid}$  & \textbf{35.2} & \textbf{16.0} & 72.3 & \textbf{82.2} & \textbf{42.3} & \textbf{44.4} & \textbf{48.5} \\
\hline
\end{tabular}
}
\caption{Main benchmark results on \textbf{Qwen2.5-Math-7B}. All values are accuracies multiplied by 100. Best results are in \textbf{bold}.}
\label{tab_exp_1}
\end{table*}

% Eurus-PRIME 只开源了Qwen2.5-Math-7B的模型

\begin{table*}[h]
\centering
{\setlength{\tabcolsep}{3pt}
\renewcommand{\arraystretch}{1.2}
\begin{tabular}{lccccccc}
\hline
\textbf{Model} & \textbf{AIME24} & \textbf{AIME25} & \textbf{AMC23} & \textbf{MATH500} & \textbf{Minerva} & \textbf{Olympiad} & \textbf{Avg.} \\
\hline
Qwen2.5-7B      & 5.1 & 2.5 & 27.8 & 34.4 & 5.9 & 13.5 & 14.9 \\
\hline
\multicolumn{8}{c}{\textbf{Classical RLVR Methods}} \\
\hline
GRPO            & 15.6 & 8.8 & 62.5 & 78.2 & 38.6 & 40.4 & 40.7 \\
\rebuttal{SimpleRL-Zoo}    & \rebuttal{17.0} & \rebuttal{9.6} & \rebuttal{64.7} & \rebuttal{76.6} & \rebuttal{31.6} & \rebuttal{40.3} & \rebuttal{40.0} \\
OPO             & 16.6 & 8.4 & 64.6 & 74.6 & 31.6 & 40.3 & 39.4 \\
\hline
\multicolumn{8}{c}{\textbf{Scheduling Methods}} \\
\hline
$\text{ACC}_{sigmoid}$             & 16.7 & 9.8 & 68.6 & 79.0 & 34.2 & 39.4 & 41.3 \\
LPPO       & 15.8 & 9.4 & 64.0 & 76.8 & 35.3 & 36.7 & 39.7 \\
Seed-GRPO       & 13.3 & 6.0 & 63.3 & 76.6 & 32.4 & 36.3 & 38.0 \\
\hline
\multicolumn{8}{c}{\textbf{Our Methods}} \\
\hline
\rowcolor{gray!10} $\text{Re-Schedule}_{linear}$    & \textbf{18.4} & 12.2 & 68.6 & 80.4 & 41.2 & 42.1 & 43.8 \\
\rowcolor{gray!10} $\text{Re-Schedule}_{sigmoid}$    & 18.2 & \textbf{14.0} & \textbf{69.2} & \textbf{81.0} & \textbf{41.5} & \textbf{43.3} & \textbf{44.5} \\
\hline
\end{tabular}
}
\caption{Main benchmark results on \textbf{Qwen2.5-7B}. All values are accuracies multiplied by 100. Best results are in \textbf{bold}.}
\label{tab_exp_2}
\end{table*}

As shown in Tables \ref{tab_exp_1} and \ref{tab_exp_2}, our Re-Schedule method consistently sets a new state-of-the-art, achieving average scores of 48.5 on Qwen2.5-Math-7B and 44.5 on Qwen2.5-7B. It significantly outperforms both scheduling baselines like $\text{ACC}_{sigmoid}$ (by up to 3.2\%) and classical RLVR methods like OPO/GRPO (by up to 3.8\%). These results validate our central claim: that the reasoning tree's structure, captured by our r-score, is a more effective way to measure the real learning difficulty of a query than path-based metrics like accuracy. 
% In summary, the main experiments provide robust evidence for the superiority of Re-Schedule. It not only sets a new state-of-the-art but also validates that a curriculum guided by the structural properties of the reasoning tree is a more powerful and principled foundation for RLVR than existing methods.

\subsection{Ablation Experiment}

% We conduct a series of ablation studies to analyze the key components of Re-Schedule: the tree construction parameters, the weight function hyperparameters, and the r-score calculation metric.

% \subsubsection{Effect of Tree Construction Parameters}

\begin{table*}[h]
\centering
{\setlength{\tabcolsep}{3pt}
\renewcommand{\arraystretch}{1.4}
\begin{tabular}{cc|ccccccc}
\hline
\textbf{Branch $k$} & \textbf{Depth $d$} & \ \textbf{AIME24} & \textbf{AIME25} & \textbf{AMC23} & \textbf{MATH500} & \textbf{Minerva} & \textbf{Olympiad} & \textbf{Avg.} \\
\hline
4 & 4 & 34.2 & 16.0 & 71.1 & 81.8 & 42.3 & 44.4 & \textbf{48.3} \\
3 & 5 & 33.8 & 14.8 & 68.4 & 79.6 & 42.3 & 42.8 & 46.9 \\
5 & 3 & 31.7 & 14.2 & 70.4 & 81.0 & 41.9 & 43.0 & 47.0 \\
\hline
\end{tabular}
}
\caption{Ablation study on tree construction parameters. The default configuration (branching factor $k=4$, depth $d=4$) achieves the best performance.}
\label{exp_ab_1}
\end{table*}

% We investigate the impact of the reasoning tree's structure by varying the branching factor ($k$) and maximum depth ($d$). As shown in Table \ref{exp_ab_1}, our default configuration of $k=4$ and $d=4$ yields the best average performance (48.3). 
% d，k的取值反应了对推理树的近似程度。d、k越大，树近似的越好，r-score的效果越好。但在实际使用中，构建过大的树会造成巨大的计算开销。因此本文采用$k=4$ and $d=4$的设置
We investigate the impact of the reasoning tree's structure by varying the branching factor $k$ and maximum depth $d$. The choice of these parameters determines the fidelity of the approximated reasoning tree. While larger values for $k$ and $d$ theoretically provide a more accurate approximation and thus a more effective r-score, they also introduce a significant computational overhead. As shown in Table \ref{exp_ab_1}, our default configuration of $k=4$ and $d=4$ yields the best average performance (48.3\%).
\rebuttal{
For more detailed analysis, please see Appendices \ref{ap_l} and \ref{ap_k_d}.
}
% , indicating that it strikes an optimal trade-off between approximation quality and computational feasibility.

% \subsubsection{Effect of Weight Function Parameters}

\begin{table*}[h]
\centering
{\setlength{\tabcolsep}{3pt}
\renewcommand{\arraystretch}{1.4}
\begin{tabular}{cc|ccccccc}
\hline
$\omega_{\text{min}}$ & $\omega_{\text{max}}$ & \ \textbf{AIME24} & \textbf{AIME25} & \textbf{AMC23} & \textbf{MATH500} & \textbf{Minerva} & \textbf{Olympiad} & \textbf{Avg.} \\
\hline
% sigmoid\_min\_0.5\_max\_2.0\_T\_10  & 35.2 & 16.0 & 72.3 & 82.2 & 42.3 & 44.4 & 48.5 \\
0.5 & 2.0  & 35.2 & \textbf{16.0} & \textbf{72.3} & \textbf{82.2} & \textbf{42.3} & \textbf{44.4} & \textbf{48.5} \\
0.5 & 1.5  & 31.4 & 15.4 & \textbf{72.3} & 81.8 & 38.1 & 42.5 & 46.9 \\
0.5 & 3.0  & 33.5 & 15.0 & 69.1 & 81.8 & 37.5 & 41.0 & 46.3 \\
0.8 & 2.0  & \textbf{36.6} & 13.6 & 71.1 & 81.6 & 37.1 & 43.8 & 47.3 \\
0.2 & 2.0  & 33.5 & 13.9 & 71.0 & 80.0 & 38.2 & 41.6 & 46.4 \\
\hline
\end{tabular}
}
\caption{Ablation study on the weight function hyperparameters, $\omega_{\text{min}}$ and $\omega_{\text{max}}$. The default setting (0.5, 2.0) performs best.}
\label{exp_ab_2}
\end{table*}

% 此外关于改变r score计算的设计选择的补充实验可见附录A

We analyze the sensitivity of our method to the minimum $\omega_{\text{min}}$ and maximum $\omega_{\text{max}}$ weight hyperparameters, which control the dynamic range of the curriculum. Results in Table \ref{exp_ab_2} show that our default setting of $\omega_{\text{min}}=0.5$ and $\omega_{\text{max}}=2.0$ achieves the highest average score (48.5). Decreasing the dynamic range by either reducing $\omega_{\text{max}}$ (to 1.5) or increasing $\omega_{\text{min}}$ (to 0.8) leads to performance degradation. This indicates that a sufficiently large weighting range is crucial for the curriculum to effectively differentiate between easy and hard samples. 
% Conversely, an overly extreme range (e.g., $\omega_{\text{min}}=0.2$) also hurts performance, possibly by starving difficult samples of training signal for too long. 
Conversely, an overly extreme range (e.g., $\omega_{\text{min}}=0.2$) also degrades performance, possibly because the curriculum excessively under-weights difficult queries. By assigning them a minimal weight for a prolonged period, the model is prevented from learning difficult queries.
Furthermore, for additional experiments on the design choices for the r-score calculation, please see Appendix \ref{ap_exp_metric}.

\begin{table}[h]
\centering
\caption{Computational cost vs. Performance gain. ``Additional Cost" is relative to the total training time.}
\label{tab:cost_analysis}
\begin{tabular}{l|cccc}
\toprule
\textbf{Tree Size} ($k^d$) & $\mathbf{3^3}$ & $\mathbf{4^3}$ & $\mathbf{3^4}$ & $\mathbf{4^4}$ (Default) \\
\midrule
Time Cost (hours) & 3.48 & 6.21 & 6.70 & 22.67 \\
Additional Cost & +7.45\% & +13.30\% & +14.35\% & +48.54\% \\
Avg Performance Gain & +3.2 & +3.0 & +3.2 & +4.0 \\
\bottomrule
\end{tabular}
\end{table}

\subsection{\rebuttal{Analysis Experiments}}

\begin{wrapfigure}{r}{0.5\textwidth} % r 表示图片在右侧，0.5\textwidth 表示图片占页面宽度的一半
\vskip -0.94in
    \centering
    \includegraphics[width=\linewidth]{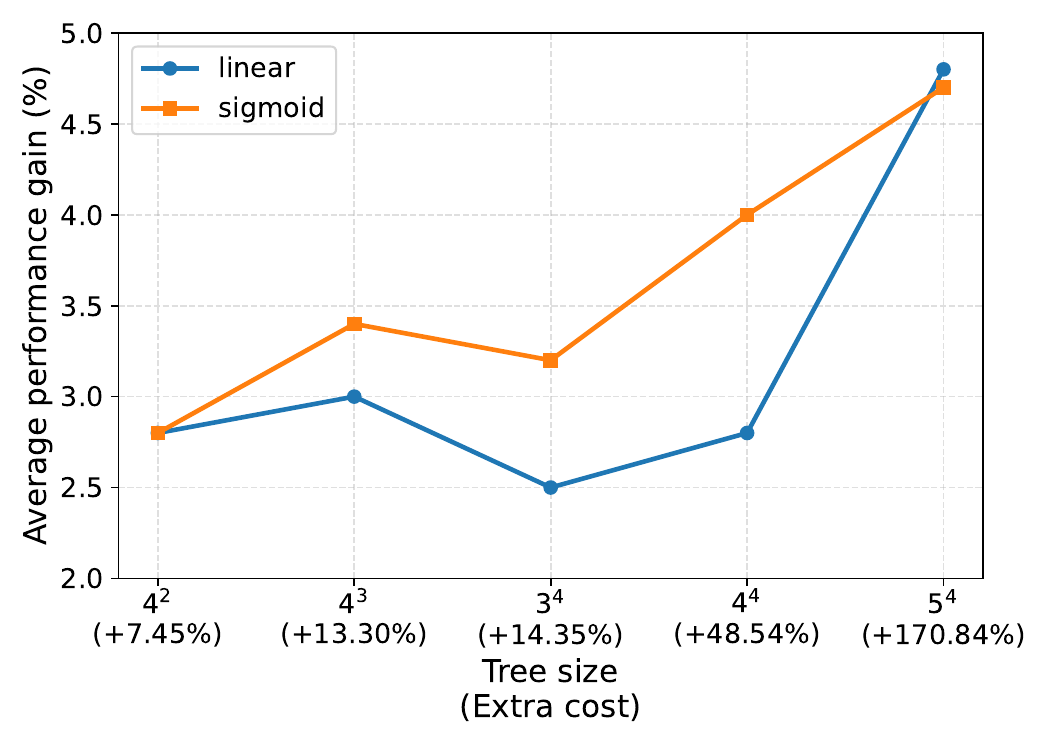}
        \vskip -0.1in
    \caption{Average performance gain versus reasoning tree size and computational cost.}
    \label{fg_trade}
    \vskip -0.1in
\end{wrapfigure}

\subsubsection{\rebuttal{Computational Cost Analysis}}
\label{app:cost_analysis}

\rebuttal{
As shown in Figure~\ref{fg_trade}, we analyzed the trade-off between the offline tree construction cost and the resulting performance gain. 
}

\rebuttal{
Table \ref{tab:cost_analysis} presents the time cost measured on 8 $\times$ H20 GPUs. While larger trees ($4^4$) incur higher preprocessing costs compared to smaller trees ($3^3$), the cost remains manageable relative to the total training time (approx. 46 hours for 5 epochs), and the performance gains are substantial.
}

\subsubsection{\rebuttal{Impact of Ordering}}
\label{app:reverse_schedule}

\begin{table}[H]
\centering
\caption{Comparison between Re-Schedule (Easy-to-Hard) and Reverse Schedule (Hard-to-Easy).}
\label{tab:reverse_schedule}
\resizebox{\textwidth}{!}{%
\begin{tabular}{l|cccccc|c}
\toprule
\textbf{Schedule} & \textbf{AIME24} & \textbf{AIME25} & \textbf{AMC23} & \textbf{MATH500} & \textbf{Minerva} & \textbf{Olympiad} & \textbf{Avg} \\
\midrule
\multicolumn{8}{c}{\textit{Linear Mapping}} \\
Re-Schedule (Ours) & 34.2 & 15.6 & 72.4 & 81.2 & 36.4 & 42.5 & {47.1} \\
Reverse Schedule & 31.9 & 14.0 & 67.6 & 81.0 & 37.8 & 41.8 & 45.7 \\
\midrule
\multicolumn{8}{c}{\textit{Sigmoid Mapping}} \\
Re-Schedule (Ours) & 34.2 & 16.0 & 71.1 & 81.8 & 42.3 & 44.4 & {48.3} \\
Reverse Schedule & 30.2 & 15.4 & 67.1 & 80.6 & 34.9 & 40.2 & 44.7 \\
\bottomrule
\end{tabular}%
}
\end{table}

\rebuttal{
To validate the ``easy-to-hard" curriculum design, we compared our method against a ``Reverse Schedule" where lower r-score (harder) samples are prioritized first. As shown in Table \ref{tab:reverse_schedule}, the Reverse Schedule leads to a significant drop in performance, confirming that starting with structurally simple samples is crucial for effective learning.
}

\section{Conclusions}

In this work, we challenged the reliance on path-based metrics for RLVR data scheduling. We introduced the r-score, a novel metric that quantifies learnability based on the structure of a query's reasoning tree, and proposed Re-Schedule, a curriculum learning algorithm built upon it. Extensive experiments demonstrated that Re-Schedule consistently outperforms classical RLVR and existing scheduling methods, validating that r-score is a more effective proxy for learnability than path-based accuracy. Our findings establish that a structural understanding of the reasoning process provides a more powerful and principled foundation for creating efficient training curricula in RLVR.

\clearpage

\newpage

\section*{Ethics statement}

We have manually reevaluated the dataset we created to ensure it is free of any potential for discrimination, human rights violations, bias, exploitation, and any other ethical concerns.

\section*{Reproducibility statement}

To ensure the reproducibility of our findings, all source code and datasets used in our experiments are included in the supplementary material. The provided materials are sufficient to replicate the main results presented in this paper.

\bibliography{iclr2026_conference}
\bibliographystyle{iclr2026_conference}

\newpage

\appendix

\section{Usage of LLMs}
\label{sec:llm_usage}

Throughout the preparation of this manuscript, Large Language Models (LLMs) were utilized as a writing and editing tool. Specifically, we employed LLMs to improve the clarity and readability of the text, refine sentence structures, and correct grammatical errors. All final content, including the core scientific claims, experimental design, and conclusions, was conceived and written by us, and we take full responsibility for the final version of this paper.

\section{Acknowledgements}
The authors would like to thank all the anonymous reviewers for their insightful comments and valuable suggestions. 
This research is supported by Anhui Provincial Natural Science Foundation (Grant No.2408085QF214), the Fundamental Research Funds for the Central Universities (Grant No.WK2080000206) and the Opening Project of the State Key Laboratory of General Artificial Intelligence (Project No.SKLAGI2025OP06). The work is also sponsored by the CodeBuddy (https://www.codebuddy.ai/). Special thanks go to Yi Liu and the team at Tencent for their generous support and assistance.

\section{Details of Experimental Setup}\label{exp_main}

% \subsection{Training Setup}\label{ap_train_set}

All algorithms are implemented based on the official GRPO codebase within the VeRL framework.
We use a learning rate of 1e-6 without warm-up across all experiments.
At each rollout step, we generate $8$ answers for each of $512$ sampled questions, then split the data into $16$ mini-batches and train the policy network for 16 gradient steps.
Models are trained for at most $150$ rollout steps.
Unless otherwise specified, we follow GRPO’s default design choices with token-level loss normalization without dynamic sampling and KL regularization.
For all models, the maximum input length is 1024 and the minimum input length is 3072.
All the experiments were conducted on H20 GPUs.

Note: The authors of Eurus-PRIME only published results from training on Qwen2.5-Math-7B. Therefore, we do not include results for the Qwen2.5-7B model in our comparison.

\section{Supplementary Experiment}

\subsection{Effect of Metric Selection}\label{ap_exp_metric}

% \subsection{Effect of Metric Selection}\label{ap_exp_metric}

\begin{table*}[h]
\centering
{\setlength{\tabcolsep}{3pt}
\renewcommand{\arraystretch}{1.2}
\begin{tabular}{lccccccc}
\hline
\textbf{Model} & \textbf{AIME24} & \textbf{AIME25} & \textbf{AMC23} & \textbf{MATH500} & \textbf{Minerva} & \textbf{Olympiad} & \textbf{Avg.} \\
\hline
Fix \\
\hline
% linear\_min\_0.5\_max\_2.0\_T\_10   & 34.2 & 15.6 & 72.4 & 81.2 & 36.4 & 42.5 & 47.1 \\
% sigmoid\_min\_0.5\_max\_2.0\_T\_10  & 34.2 & 16.0 & 71.1 & 81.8 & 42.3 & 44.4 & 48.3 \\
$\text{Re-Schedule}_{linear}$     & 34.2 & 15.6 & 72.4 & 81.2 & 36.4 & 42.5 & 47.1 \\
$\text{Re-Schedule}_{sigmoid}$  & 34.2 & 16.0 & 71.1 & 81.8 & 42.3 & 44.4 & 48.3 \\
\hline
Pruning \\
\hline
$\text{Re-Schedule}_{linear}$    & 35.7 & 14.6 & 73.7 & 81.0 & 34.9 & 41.6 & 46.9 \\
$\text{Re-Schedule}_{sigmoid}$  & 33.1 & 16.7 & 71.1 & 82.0 & 39.0 & 42.4 & 47.4 \\
\hline
\end{tabular}
}
\caption{Ablation study comparing our proposed node-level modification metric (`Fix') with a branch-level `Pruning' metric. `Fix' consistently outperforms `Pruning', validating our fine-grained approach.}
\label{exp_ab_3}
\end{table*}

We validate our core design choice for the r-score calculation. Our proposed method (`Fix') defines an `edit' as a single node modification. We compare this against an alternative (`Pruning'), where an `edit' is defined as pruning an entire sub-branch from a decision point. 
Table \ref{exp_ab_3} shows that the `Fix' method consistently outperforms `Pruning' for both linear (47.1\% vs. 46.9\%) and sigmoid (48.3\% vs. 47.4\%) schedules. 
% This result confirms our hypothesis that a fine-grained, node-level analysis provides a more accurate measure of a query's learning potential. The `Pruning` method is too coarse, as it fails to distinguish between correcting a single critical node and removing an entire subtree, thereby losing crucial structural information that our `Fix` method effectively captures.
This result shows that compared with the branch `Pruning', the node `Fix' is more consistent with the training process of reinforcement learning.

\subsection{\rebuttal{Variance Analysis}}

\begin{table}[h]
\centering
\caption{Variance of Re-Schedule over multiple runs.}
\label{tab:stability_runs}
\resizebox{\textwidth}{!}{%
\begin{tabular}{l|cccccc}
\toprule
 & \textbf{AIME24} & \textbf{AIME25} & \textbf{AMC23} & \textbf{MATH500} & \textbf{Minerva} & \textbf{Olympiad} \\
\midrule
Avg. $\pm$ Var.  & 35.2 $\pm$ 0.1 & 16.0 $\pm$ 0.0 & 72.3 $\pm$ 0.4 & 82.2 $\pm$ 0.0 & 42.3 $\pm$ 0.6 & 44.4 $\pm$ 0.6  \\
\bottomrule
\end{tabular}%
}
\end{table}

\rebuttal{
To assess the stability of our proposed Re-Schedule method, we conducted repeated experiments using different random seeds. Table \ref{tab:stability_runs} reports the variance. The results demonstrate that our method exhibits low variance. This confirms that the reported improvements are statistically stable and not due to random variation. Compared to performance improvements, the impact of variance is minimal. 
}

\subsection{\rebuttal{Ablation on Token Interval $l$ }}\label{ap_l}

\rebuttal{
To investigate the impact of $l$ on performance and its relationship with model capabilities, we conducted ablation studies on two base models: {Qwen2.5-Math-7B} and {Qwen2.5-7B}. We utilized the Sigmoid weighting mapping for these experiments.
}

\rebuttal{
Table \ref{tab:ablation_l_summary} summarizes the average accuracy across six benchmarks (AIME24, AIME25, AMC23, MATH500, Minerva, Olympiad) for varying token intervals $l \in \{200, 400, 600, 1200\}$.
}

\begin{table}[h]
\centering
\caption{Impact of Token Interval $l$ on Average Accuracy (Sigmoid Mapping).}
\label{tab:ablation_l_summary}
\begin{tabular}{l|cc}
\toprule
\textbf{Interval ($l$)} & \textbf{Qwen2.5-Math-7B (Avg)} & \textbf{Qwen2.5-7B (Avg)} \\
\midrule
$l=200$  & 48.3 & {44.5} \\
$l=400$  & 48.6 & 43.2 \\
$l=600$  & {48.9} & 43.1 \\
$l=1200$ & 46.0 & 41.1 \\
\bottomrule
\end{tabular}
\end{table}

\rebuttal{
The results indicate that Re-Schedule is generally robust to the choice of $l$ within the range of $[200, 600]$.
For the specialized math model (Qwen2.5-Math-7B), performance remains high and stable as $l$ increases to 600. For the general model (Qwen2.5-7B), while $l=200$ yields the best results, the performance drop at $l=600$ is relatively contained.
A significant performance drop is observed for both models when $l=1200$. This suggests that when the interval is too large, the approximated reasoning tree becomes too coarse to capture the critical branching points necessary for effective r-score estimation.
}

% \subsection{\rebuttal{Fair Comparison with SimpleRL-Zoo}}

% \begin{table}[h]
% \centering
% \caption{Comparison between SimpleRL-Zoo (reproduced on DAPO) and Re-Schedule.}
% \label{tab:fair_comparison}
% \resizebox{\textwidth}{!}{%
% \begin{tabular}{l|cccccc|c}
% \toprule
% \textbf{Model} & \textbf{AIME24} & \textbf{AIME25} & \textbf{AMC23} & \textbf{MATH500} & \textbf{Minerva} & \textbf{Olympiad} & \textbf{Avg} \\
% \midrule
% SimpleRL-Zoo (Published) & 25.2 & 13.4 & 70.6 & 78.6 & 37.9 & 38.4 & 44.0 \\
% SimpleRL-Zoo (DAPO-Trained) & 30.8 & 14.2 & 65.4 & 79.2 & 37.1 & 40.8 & 44.6 \\
% $\text{Re-Schedule}_{linear}$ & {34.2} & {15.6} & {72.4} & {81.2} & {36.4} & {42.5} & {47.1} \\
% $\text{Re-Schedule}_{sigmoid}$ & {35.2} & {16.0} & {72.3} & {82.2} & {42.3} & {44.4} & {48.5} \\
% \bottomrule
% \end{tabular}%
% }
% \end{table}

% \rebuttal{
% To ensure a fair comparison, we reproduced the SimpleRL-Zoo baseline using the exact same training set used in our work (DAPO-Math-17k), as opposed to the original paper's use of GSM8k and MATH. Table \ref{tab:fair_comparison} shows that even when trained on identical data, Re-Schedule consistently outperforms the SimpleRL-Zoo baseline.
% }

\subsection{\rebuttal{Sensitivity to Branching Factor $k$, Depth $d$ and Modification Budget $M$}}\label{ap_k_d}

\rebuttal{
We investigated the impact of the reasoning tree size on performance by varying the branching factor $k$ and depth $d$ on the Qwen2.5-Math-7B model. 
}

\begin{table}[h]
\centering
\caption{Ablation study on branching factor $k$ (with fixed $d=4$). }
\label{tab:ablation_k}
\resizebox{\textwidth}{!}{%
\begin{tabular}{l|cccccc|c}
\toprule
\textbf{Setting} & \textbf{AIME24} & \textbf{AIME25} & \textbf{AMC23} & \textbf{MATH500} & \textbf{Minerva} & \textbf{Olympiad} & \textbf{Avg} \\
\midrule
\multicolumn{8}{c}{\textit{Linear Mapping}} \\
$k=3$ & 32.5 & 15.2 & 74.0 & 81.8 & 36.3 & 41.0 & 46.8 \\
$k=4$ (Default) & 34.2 & 15.6 & 72.4 & 81.2 & 36.4 & 42.5 & 47.1 \\
$k=5$ & 35.1 & 18.1 & 77.4 & 81.8 & 37.8 & 44.6 & {49.1} \\
\midrule
\multicolumn{8}{c}{\textit{Sigmoid Mapping}} \\
$k=3$ & 32.0 & 15.2 & 74.2 & 81.6 & 38.6 & 43.1 & 47.5 \\
$k=4$ (Default) & 34.2 & 16.0 & 71.1 & 81.8 & 42.3 & 44.4 & 48.3 \\
$k=5$ & 36.4 & 17.0 & 75.5 & 81.2 & 40.6 & 43.4 & {49.0} \\
\bottomrule
\end{tabular}%
}
\end{table}

\begin{table}[h]
\centering
\caption{Ablation study on tree depth $d$ (with fixed $k=4$). }
\label{tab:ablation_d}
\resizebox{\textwidth}{!}{%
\begin{tabular}{l|cccccc|c}
\toprule
\textbf{Setting} & \textbf{AIME24} & \textbf{AIME25} & \textbf{AMC23} & \textbf{MATH500} & \textbf{Minerva} & \textbf{Olympiad} & \textbf{Avg} \\
\midrule
\multicolumn{8}{c}{\textit{Linear Mapping}} \\
$d=2$ & 31.1 & 15.3 & 74.2 & 81.8 & 38.2 & 42.2 & 47.1 \\
$d=3$ & 31.4 & 14.6 & 72.7 & 82.0 & 39.7 & 43.3 & 47.3 \\
$d=4$ (Default) & 34.2 & 15.6 & 72.4 & 81.2 & 36.4 & 42.5 & {47.1} \\
\midrule
\multicolumn{8}{c}{\textit{Sigmoid Mapping}} \\
$d=2$ & 31.9 & 14.8 & 74.5 & 81.6 & 37.2 & 42.4 & 47.1 \\
$d=3$ & 33.2 & 16.4 & 73.0 & 80.0 & 41.9 & 41.6 & 47.7 \\
$d=4$ (Default) & 34.2 & 16.0 & 71.1 & 81.8 & 42.3 & 44.4 & {48.3} \\
\bottomrule
\end{tabular}%
}
\end{table}

% --- Table for M (New) ---
\begin{table}[h]
\centering
\caption{Ablation study on node modification budget $M$ (with fixed $k=4, d=4, l=200$).}
\label{tab:ablation_m}
\resizebox{\textwidth}{!}{%
\begin{tabular}{l|cccccc|c}
\toprule
\textbf{Setting} & \textbf{AIME24} & \textbf{AIME25} & \textbf{AMC23} & \textbf{MATH500} & \textbf{Minerva} & \textbf{Olympiad} & \textbf{Avg} \\
\midrule
$M=5$ & 33.6 & 15.4 & 72.2 & 79.0 & 40.4 & 43.1 & 47.3 \\
$M=10$ (Default) & 34.2 & 16.0 & 71.1 & 81.8 & 42.3 & 44.4 & \textbf{48.3} \\
$M=15$ & 34.7 & 16.0 & 71.8 & 82.0 & 41.6 & 42.4 & 48.1 \\
\bottomrule
\end{tabular}%
}
\end{table}

\rebuttal{
\textbf{Varying branching factor $k$:} Fixing $d=4$ and $l=200$, we tested $k \in \{3, 4, 5\}$. As shown in Table \ref{tab:ablation_k}, increasing $k$ generally improves performance, suggesting that a denser tree captures the structural difficulty more accurately.
}

\rebuttal{
\textbf{Varying tree depth $d$:} Fixing $k=4$ and $l=200$, we tested $d \in \{2, 3, 4\}$. Table \ref{tab:ablation_d} shows that deeper trees provide a better estimation of the reasoning structure, leading to improved downstream performance.
}

\rebuttal{
\textbf{Varying node modification budget $M$:} Finally, we assess the stability of our method with respect to the node modification budget $M$. Fixing $k=4, d=4$, and $l=200$, we evaluated performance across $M \in \{5, 10, 15\}$. As presented in Table \ref{tab:ablation_m}, the results are relatively robust to changes in this parameter. While the default setting of $M=10$ yields the optimal average accuracy, varying the budget between 5 and 15 results in no significant performance degradation, indicating that the r-score remains a reliable metric across different budget constraints.
}

\subsection{\rebuttal{Generalization to Different Model Architectures}}
\label{app:qwen3_generalization}

\rebuttal{
To demonstrate the broad applicability of our method beyond the Qwen2.5 family, we conducted additional experiments on the {Qwen3-4B-Base} model. 
}

\begin{table}[h]
\centering
\caption{Performance comparison on {Qwen3-4B-Base}.}
\label{tab:qwen3_results}
\resizebox{\textwidth}{!}{%
\begin{tabular}{l|cccccc|c}
\toprule
\textbf{Model} & \textbf{AIME24} & \textbf{AIME25} & \textbf{AMC23} & \textbf{MATH500} & \textbf{Minerva} & \textbf{Olympiad} & \textbf{Avg} \\
\midrule
GRPO & 24.2 & 21.8 & 52.4 & 86.0 & 39.4 & 43.4 & 44.5 \\
ACC & 24.8 & 23.3 & 59.3 & 88.8 & 41.6 & 42.0 & 46.6 \\
{Re-Schedule (Ours)} & {27.6} & {26.9} & {57.6} & {89.8} & {43.5} & {47.4} & {48.8} \\
\bottomrule
\end{tabular}%
}
\end{table}

\rebuttal{
As shown in Table \ref{tab:qwen3_results}, Re-Schedule consistently outperforms both the standard GRPO baseline and the accuracy-based curriculum (ACC) across all benchmarks. This confirms that the effectiveness of the r-score is not limited to specific model architectures or sizes.
}

\newpage

\subsection{\rebuttal{Dynamic R-Score Calculation}}
\label{app:dynamic_rscore}

\rebuttal{
To determine if the r-score should be updated as the model evolves, we compared our standard static approach (computed once before training) with a dynamic approach where the r-score is re-computed and weights are updated three times during the training process.
}

\begin{table}[h]
\centering
\caption{Comparison between Static and Dynamic R-Score updates on Qwen2.5-Math-7B.}
\label{tab:dynamic_rscore}
\resizebox{\textwidth}{!}{%
\begin{tabular}{l|cccccc|c}
\toprule
\textbf{Method} & \textbf{AIME24} & \textbf{AIME25} & \textbf{AMC23} & \textbf{MATH500} & \textbf{Minerva} & \textbf{Olympiad} & \textbf{Avg} \\
\midrule
\multicolumn{8}{c}{\textit{Linear Mapping}} \\
Static (Default) & 34.2 & 15.6 & 72.4 & 81.2 & 36.4 & 42.5 & 47.1 \\
Dynamic (3 updates) & 34.6 & 14.9 & 75.3 & 80.0 & 39.7 & 41.6 & 47.8 \\
\midrule
\multicolumn{8}{c}{\textit{Sigmoid Mapping}} \\
Static (Default) & 34.2 & 16.0 & 71.1 & 81.8 & 42.3 & 44.4 & 48.3 \\
Dynamic (3 updates) & 35.3 & 15.2 & 74.2 & 82.4 & 42.7 & 43.6 & 48.9 \\
\bottomrule
\end{tabular}%
}
\end{table}

\rebuttal{
As shown in Table \ref{tab:dynamic_rscore}, the dynamic approach yields performance comparable to the static baseline (e.g., 48.9\% vs. 48.3\% for Sigmoid mapping). Given the substantial computational cost of re-generating reasoning trees during training, we conclude that the static r-score serves as a sufficient and efficient prior for guiding the curriculum.
}

\newpage

\subsection{\rebuttal{Generalization to Code Generation}}
\label{app:code_gen}

\begin{table}[h]
\centering
\caption{Performance comparison on Code Generation (LiveCodeBench v5).}
\label{tab:code_gen_results}
\begin{tabular}{l|cc}
\toprule
\textbf{Method} & \textbf{pass@1} & \textbf{pass@4} \\
\midrule
GRPO & 25.4 & 35.4 \\
$\text{ACC}_{sigmoid}$ & 25.8 & 36.0 \\
$\text{Re-Schedule}_{sigmoid}$ & {26.3} & {37.8} \\
\bottomrule
\end{tabular}
\end{table}

\rebuttal{
To validate the generalization capability of Re-Schedule beyond mathematical reasoning, we extended our evaluation to the domain of Code Generation.
We utilized {DeepSeek-R1-Distill-Qwen-1.5B} as the base model. The model was trained on the {ArcherCodeR} dataset~\cite{wang2025stabilizing}, which contains 6,753 code generation tasks. For evaluation, we used the {LiveCodeBench v5} benchmark~\cite{jain2024livecodebench}. We report {pass@1} and {pass@4} metrics (averaged over 8 samples).
}

\rebuttal{
As shown in Table \ref{tab:code_gen_results}, Re-Schedule consistently outperforms both the standard GRPO baseline and the accuracy-based curriculum (ACC). Specifically, our method achieves a {+0.9\%} improvement in pass@1 and a significant {+2.4\%} improvement in pass@4 compared to GRPO. These results confirm that the structural insights captured by the r-score are effective in the coding domain, where the ``reasoning tree'' corresponds to the decision space of code logic and syntax.
Looking forward, the tree-based structural metrics could also be adapted to improve multi-modal data generation pipelines and self-instructed compositional code captioning \citep{wang2024world, wang2025grasp, wang2025vgr, hao2026recreate, wang2025traceable, lei2025scalability}.
}

\end{document}